\documentclass{article}
\usepackage[utf8]{inputenc}
\usepackage{graphicx}
\graphicspath{{Images/}}

    \PassOptionsToPackage{numbers, compress}{natbib}
    \usepackage[preprint]{neurips_2022}

\usepackage{amsmath}
\usepackage{amssymb}
\usepackage[utf8]{inputenc} % allow utf-8 input
\usepackage[T1]{fontenc}    % use 8-bit T1 fonts
\usepackage{hyperref}       % hyperlinks
\usepackage{url}            % simple URL typesetting
\usepackage{booktabs}       % professional-quality tables
\usepackage{amsfonts}       % blackboard math symbols
\usepackage{nicefrac}       % compact symbols for 1/2, etc.
\usepackage{microtype}      % microtypography
\usepackage{xcolor}         % colors
\usepackage{graphicx}
\usepackage{booktabs}
\usepackage{multirow}
\usepackage{mathrsfs}

\title{The Short Text Matching Model Enhanced with Knowledge via Contrastive Learning}
\author{
Ruiqiang Liu\thanks{These authors contributed to the work equllly and should be regarded as co-first authors.} \\
  Alibaba \\
\texttt{liuruiqiang.lrq@alibaba-inc.com} \\
\And
Qiqiang Zhong\footnotemark[1] \\
  Xiaohongshu \\
\texttt{zhongqiqianga@163.com} \\
\And
Mengmeng  Cui \\
East China University of Science and Technology \\
\texttt{y30201052@mail.ecust.edu.cn} \\
\And
Hanjie Mai \\
Yunnan University \\
\texttt{12021215147@mail.ynu.edu.cn} \\
\And
Qiang Zhang \\
Alibaba \\
\texttt{seven.zq@alibaba-inc.com} \\
\And
Shaohua Xu \\
  Alibaba \\
\texttt{saohua.xsh@alibaba-inc.com} \\
 \And
Xiangzheng Liu \\
  Alibaba \\
\texttt{xiangzheng.lxz@alibaba-inc.com} \\
 \And
Yanlong Du \\
  Alibaba \\
\texttt{yanlong.dyl@alibaba-inc.com} \\
}

\begin{document}

\maketitle

\begin{abstract}
In recent years, short Text Matching tasks have been widely applied in the fields of advertising search and recommendation. The difficulty lies in the lack of semantic information and word ambiguity caused by the short length of the text. Previous works have introduced complement sentences or knowledge bases to provide additional feature information. However, these methods have not fully interacted between the original sentence and the complement sentence, and have not considered the noise issue that may arise from the introduction of external knowledge bases. Therefore, this paper proposes a short Text Matching model that combines contrastive learning and external knowledge. The model uses a generative model to generate corresponding complement sentences and uses the contrastive learning method to guide the model to obtain more semantically meaningful encoding of the original sentence. In addition, to avoid noise, we use keywords as the main semantics of the original sentence to retrieve corresponding knowledge words in the knowledge base, and construct a knowledge graph. The graph encoding model is used to integrate the knowledge base information into the model. Our designed model achieves state-of-the-art performance on two publicly available Chinese Text Matching datasets, demonstrating the effectiveness of our model.
\end{abstract}

\section{Introduction}
% 短文本匹配是一个......任务
% 
Short Text Matching (STM) is a task of determining semantic similarity or relevance between two short texts, such as sentence pairs, phrases or keywords. This is a crucial task in many natural language processing applications, including information retrieval, question answering, and document classification.
% 如果加进来的话就这样写
% 背景：

Search advertising is a form of advertising displayed on search engine results pages. When a user searches for a keyword in a search engine, if the keyword matches the keywords purchased by the advertiser, the advertisements will be triggered to display. Advertisements matching methods include exact match, phrase match, and broad match. Exact match and phrase match are traditional string matches, making it difficult to match text that may have slight variations in wording but have the same meaning. Broad match is usually used when there is little overlap between the text a user is searching for and the keywords purchased by the advertiser, but the meaning expressed is the same. Therefore, compared to exact and phrase matches, broad match is easier to trigger more ads due to less strict matching conditions.

With the rapid development of deep learning in recent years, vector retrieval for Text Matching has achieved significant success in multiple fields. Vector retrieval refers to the method of using mathematical representations of data to perform similarity searches in large databases. This method can be applied to various types of data, including text, images, and audio.In vector retrieval, the text searched by the user and the keywords purchased by the advertiser are first represented as high-dimensional vectors. Then, a similarity measure is used to compare these vectors to find the most relevant results for a given query. Short-text vector retrieval is the most common form of broad matching, and data statistics have shown that over 85 perenct of search content and advertiser purchased keywords in our search engine are less than 25 characters long. Therefore, short-text vector retrieval matching is more important.

However, there are many challenges in short text vector retrieval matching currently. Because short texts that users search for often lack context, there can be multiple interpretations, and limited vocabulary means that many words can have multiple meanings. As shown in the left part of Figure 1, when a user searches for "Changan," it can refer to both a car brand and a city, making it difficult to match accurately. Even though the query and the returned entity from the search results are the same, the meaning expressed in the sentence is completely different and contrary to the user's search intent. Therefore, many of the keywords purchased by keywords advertiser for matching are ineffective.

\begin{figure}[ht]
\includegraphics[width=\textwidth]{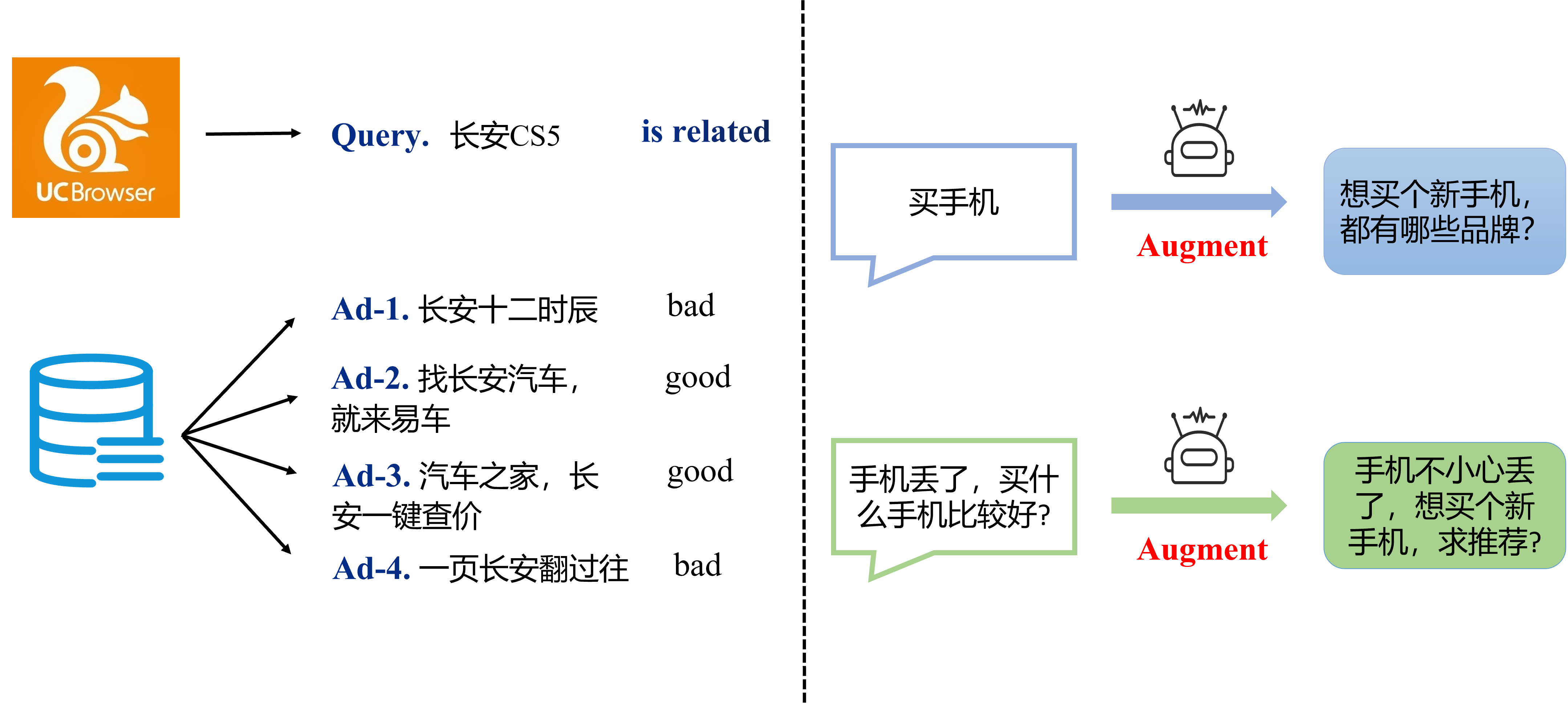}
\caption{The left image showcases bad cases in the UC Browser, while the right image demonstrates the process of generating a complement sentence.} \label{fig1}
\end{figure}
issue of entity ambiguity in short texts, previous research has often relied on incorporating externalt knowledge to supplement the semantic information of the text. For example, in Chen et al. \cite{chen2022context}, external knowledge was collected by gathtering text data from user queries and clicks in search engines, and then used to enhance the semantic information of the text. In the works of Lyu et al. \cite{lyu2021let} and Ma et al. \cite{ma2022ote}, a multi-dimensional attention mechanism was proposed to interactively fuse each word in the word grid diagram with external words retrieved from HowNet\cite{dong2003hownet}, in order to eliminate word ambiguity in the sentence.

While these methods have been proven effective through experiments, they also have certain limitations. Firstly, Chen et al. \cite{chen2022context} did not pay sufficient attention to the relationship between the original and supplemented sentences, and lacked effective interaction between the two. In fact, the original and supplemented sentences form a semantic matching sentence pair, so the supplemented sentence can guide the model to generate encoding vectors that better align with the semantics of the original sentence. In the works of Lyu et al. \cite{lyu2021let} and Ma et al. \cite{ma2022ote}, all words in the sentence were retrieved from the knowledge base, but some words are not main components of the sentence, and using them to retrieve knowledge words may introduce some noise.

Therefore, from these perspectives, we propose a short Text Matching model(KSTM) that enhances knowledge through contrastive learning. This model utilizes the SimBERT\footnote{https://github.com/ZhuiyiTechnology/simbert} text generation model to generate complement text for two input sentences. To enable SimBERT to generate text that fits the context of the original sentences, we train it using a large amount of UC and Quark browser search data from within our company. In our experiments, we use the generated complement text as positive samples, as shown on the right side of Figure 1, and other samples in the same batch as negative samples. Additionally, we extract keywords from the original sentences and use the HowNet knowledge base Dong et al. \cite{dong2003hownet} to obtain the top k knowledge words that are similar to the keywords. These words are used to construct a knowledge graph, where the nodes are keywords and knowledge words, and the edges represent the similarity weights between the keywords and similar words. The model's discriminative ability is enhanced through graph learning. Currently, our model has achieved state-of-the-art performance on two publicly available Chinese datasets.

In conclusion, we made the following contributions:
\begin{itemize}
\item We use a generative model to generate complement text for contrastive learning, in order to better guide the model to encode the original text and incorporate more contextual information.
\item We extract keywords from the original text and construct a graph by querying related words from a knowledge base, in order to eliminate entity ambiguity issues.
\item Based on the above two points, we have built a  short-Text Matching model, which achieved state-of-the-art results on two publicly available Chinese datasets, demonstrating the effectiveness of the model.

\end{itemize}

\section{Related work}
\label{gen_inst}

Short Text Matching is a widely used technology in the field of Natural Language Processing(NLP), and in recent years, a series of neural network-based Text Matching models have emerged, which can be roughly categorized as follows:

\textbf{(a) By utilizing contextual features.} Hu et al. \cite{hu2021context} proposed a context-aware interaction network (context-aware interaction network) for question matching. The interaction network in the paper is a multi-layer interaction network with a multi-head attention mechanism similar to the transformer. Each layer of the interaction network includes a context-aware cross-attention mechanism and a gate fusion layer. The former is responsible for learning the contextual interaction information when aligning the two sentences, while the latter implements flexible selection of useful alignment interaction information. Previous sentence matching models directly perform attention interaction after embedding two sentences. Hu et al. \cite{hu2020enhanced} proposed an enhanced sentence alignment network with a simple gate feature to flexibly integrate original words and contextual features to improve the attention interaction part across sentences. In short Text Matching, there is little further recognition of discriminative features and feature denoising to enhance relevance learning. Li et al. \cite{li2022adaptive} designed ADDAX to clearly distinguish distinguishing features and filter out irrelevant features in a context-aware manner.

\textbf{(b) Using token features.} Chen et al. \cite{chen2020neural} proposed a neural graph matching method (GMN) for Chinese short Text Matching. The traditional approach of segmenting each sentence into a word sequence is changed, and all possible word segmentation paths are retained to form a word lattice graph, and node representations are updated based on graph matching attention mechanism. In addition, current interactive matching methods for sentence modeling are relatively simple and ignore the importance of the relative distance between words. Deng et al. \cite{deng2022enhanced} established a high-performance distance-aware self-attention and multi-level matching model (DSSTM) for sentence semantic matching tasks. By considering the importance of tokens at different distances, better sentence semantics can be obtained.

\textbf{(c) Using additional features.} Short Text Matching focuses on learning semantic features based on character and word-level features, to some extent ignoring the special features of Chinese, such as Pinyin, radicals, and keywords. Zhao et al. \cite{zhao2022multi} proposed a multi-granularity interaction model based on Pinyin and radicals (MIPR) for Chinese semantic matching. Keywords represent factual information that needs to be strictly matched, such as actions, entities, and events, while intents convey abstract concepts and ideas that can be interpreted as various expressions. A simple yet effective training strategy was proposed in Zou et al. \cite{zou2022divide}, which involves breaking down keywords and intents and performing semantic matching on the text in a divide-and-conquer approach. Lu et al. \cite{lu2022mkpm} presented a multi-keyword matching and sentence matching method that utilizes keyword pairs from two sentences to represent their semantic relationship, thereby avoiding noise redundancy and interference that may arise from using the entire sentence.

\textbf{(d) Utilizing external information.}  Chen et al. \cite{chen2022context} obtained an external knowledge collection for each short text by using a search engine. The corresponding clicked title was used as a context sentence set to enhance the model's understanding of short texts. Lyu et al. \cite{lyu2021let} proposed using a multi-dimensional attention mechanism to enable interaction and fusion between each word in the word lattice graph and the external words retrieved from HowNet. This approach aims to eliminate ambiguity among the words in the sentence. Based on the research in Ma et al. \cite{ma2022ote}, OTE model used SoftLexicon to provide more detailed information at different levels. They used LaserTagger\cite{malmi2019encode} model and  EDA\cite{wei2019eda} to improve their data.

% All headings should be lower case (except for first word and proper nouns),
% flush left, and bold.
\section{Methodology}
\label{headings}

\begin{figure}[ht]
\includegraphics[width=\textwidth]{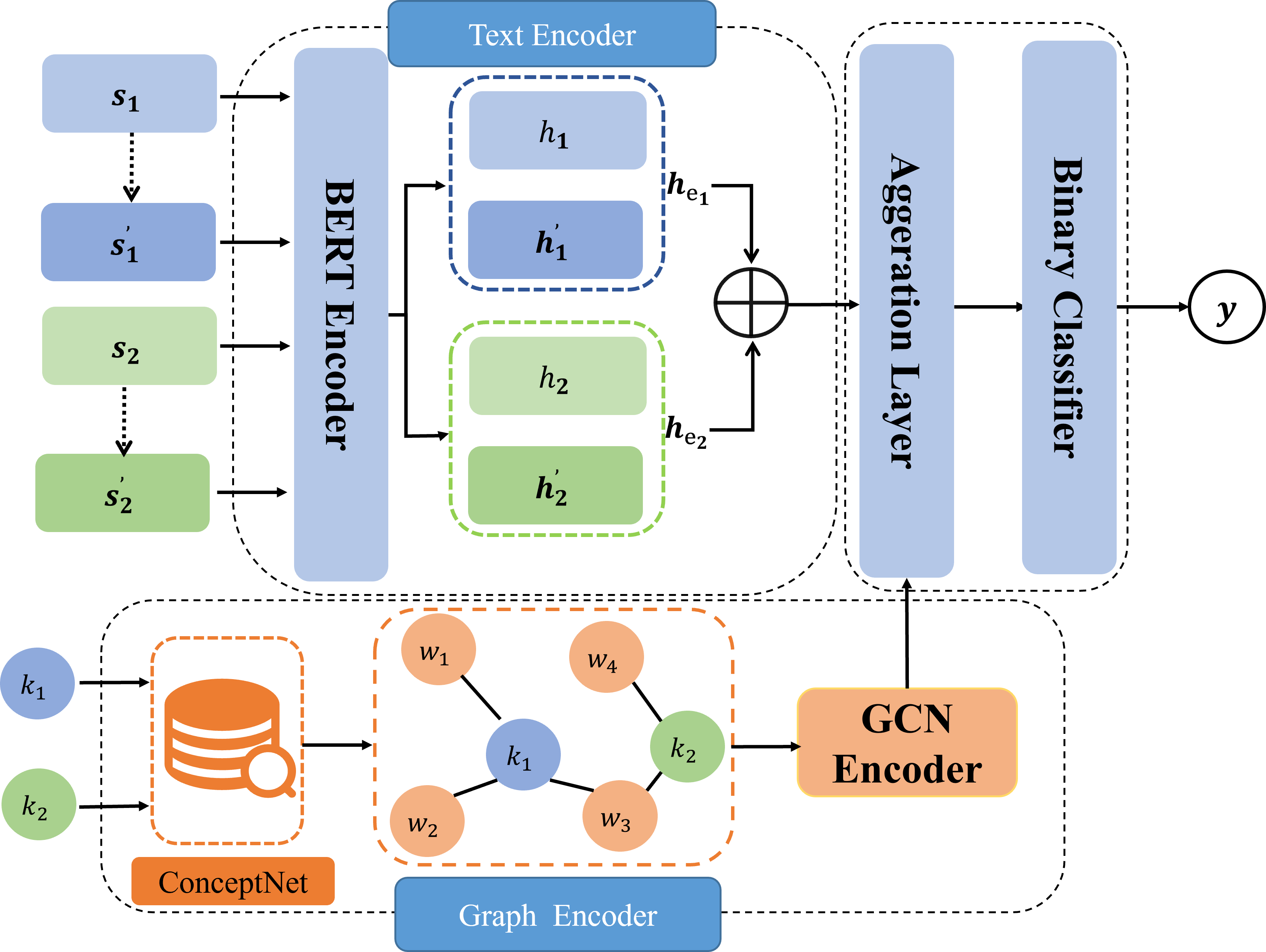}
\caption{Framework of KSTM} \label{fig3}
\end{figure}
% \vspace{-20pt}
The overview of our model (KSTM) is shown in Figure 2. It is comprised of three major components:1) Text Encoding Module, 2) Graph Encoding Module, 3) Aggeration Layer, 4)Binary Classifier Layer.

In the following section, we provide a definition of the STM task and introduce each component of the model.
% 模型框架描述：
%1. 输入句子对和数据增强句子对：模型接收两个句子对（原始句子对和经过数据增强的句子对）作为输入。
%2. 共享的BERT编码器：这两个句子对通过一个共享的BERT编码器，得到四个句子的向量表示。
%3. 关键词抽取和Hownet映射：从原始句子对中抽取关键词，并使用外部知识库Hownet进行映射，得到一个图。
%4. GCN编码器：将上一步得到的图输入到一个图卷积网络（GCN）编码器中，得到句子的表示。
%5. 表示拼接与融合：将图视角下的四个子句表示和文本视角下的四个子句表示拼接融合。
%6. 二分类预测：利用拼接融合后的表示进行二分类任务，预测句子对是否匹配。
%7. 对比学习：在表示学习的过程中，模型还采用了对比学习的方法对文本视角下的两个句子进行表示学习。
% First-level headings should be in 12-point type.
\subsection{Task Definition}
Given two original sentences $s_1$ and $s_2$, the objective of STM is to determine whether they have the same semantic meaning. It can be considered as a binary classification task. To capture more comprehensive semantic information of $s_1$ and $s_2$, we generated complement sentences $s^{'}_{1}$ and $s^{'}_{2}$ specifically for them. The details of the generation process can be found in Section 3.6.

\subsection{Text Encoding Module}

Given the remarkable performance of the pre-trained model BERT\cite{kenton2019bert} in a range of natural language processing tasks, we utilize it as the encoder to separately encode the original and complementary sentences, generating high-dimensional vectors. Specifically, we first insert the $[CLS]$ token to the beginning of each sentence, and then encode them using BERT. Finally, by extracting the representation vector of the $[CLS]$ token, we obtain the sentence representations $h_1\in\mathbb R^{1\times d}$ and $h_2\in\mathbb R^{1\times d}$ for the original sentences and $h^{'}_1\in\mathbb R^{1\times d}$ and $h^{'}_2\in\mathbb R^{1\times d}$ for the complementary sentences. $d$ is the embedded dimension. It is worth noting that reduce the parameters count  and prevent over-fitting, each sentence shares the same encoder.

Finally, we combine the semantic representations of the complementary sentence with those of the original sentence to obtain an enhanced sentence representation $h_{e1}$ and $h_{e2}$. This process improves the model's ability to discern the semantic similarity between two sentences.
\begin{equation}
h_{e1} = [h_1;h^{\prime}_1]
\end{equation}
\begin{equation}
h_{e2} = [h_2;h^{\prime}_2]
\end{equation}
\subsection{Graph Encoding Module}

The keywords of a sentence contain crucial information that the sentence intends to convey. By analyzing the relationships between keywords, the model can better infer the semantic relationship between two sentences. Therefore, to extract and represent the relationships between keywords, In this module, we utilize the keywords from the original sentence to retrieve relevant knowledge words from external knowledge sources\cite{dong2003hownet}. Subsequently, we construct a graph with these keywords and knowledge words as nodes, and the edges of the graph represent the relationships between words. Finally, we utilize the GCN\cite{kipfsemi} to extract the relationship features between words and obtain a graph representation vector.

\subsubsection{Construction Knowledge Graphs}
To construct the relationship graph, we first extract the keywords $k_1$ and $k_2$ from the two original sentences using the TextRank algorithm\cite{mihalcea2004textrank}. Specifically, $k_1$ represents the keyword of the original sentence $s_1$, while $k_2$ represents the keyword of the original sentence $s_2$. Next, we retrieve the knowledge words $W=\{w_1,w_2,..,w_n\}$ related to these keywords and their corresponding relevance scores from external knowledge sources. Finally, we consider the keywords and knowledge words as the nodes of the relationship graph, where the knowledge words related to the keywords are connected by edges with weights based on their relevance scores. The  relationship graph is defined as $G=(V, E, A)$, where $V=\{k_1,k_2,w_1,...w_n\}$ is a set of graph nodes, $E$ is the set of edges, and $A$ is the adjacency matrix of $G$.

\subsubsection{Graph Encoder}
The Graph Convolutional Network (GCN)\cite{kipfsemi}, which is a type of multi-layer neural network, can directly process data with graph structures and introduce node representation vectors based on the neighborhood features of each node. When there is only one layer in the network, GCN can only capture information from the direct neighbors of each node, and thus it requires stacking multiple layers to integrate more node information to complete the encoding of the entire graph. After constructing the relation graph, we initialize each node's encoding as $H$ and input it into a sentence encoder. Then, we feed the adjacency matrix $A$ and the node encoding $H$ into GCN for further processing.

\begin{equation}
H^{(i+1)} = Relu(\widetilde{A}H^{(i)}W^{(i)})
\end{equation}
\begin{equation}
\widetilde{A}=D^{-\frac{1}{2}}AD^{-\frac{1}{2}}
\end{equation}

where $H^{(i)}\in\mathbb R^{(n+2)\times d}$ is the representation vector of all nodes in $ith$ GCN layer, $H^(0)=H$.  $\widetilde{A}$ is the normalized symmetric adjacency matrix and $W^(i)$ is a weight matrix for $ith$ GCN layer. $D$ is degree matrix of $A$, $D_{ii}=\sum_{j}A_{ij}$.

Finally, we merge all the node representations outputted by the last layer of GCN to obtain the graph representation vector.

\begin{equation}
h_{graph} = \sum^{n}_{i=1}H^{last}_{i}
\end{equation}  
where $H^{last}$ is the representation of all nodes output by the last layer of GCN..

% Third-level headings should be in 10-point type.

\subsection{Aggregration Layer and Binary Classifier}
During the prediction process, our model combines the representations of sentences and graphs to predict the similarity scores between the original sentences.
\begin{equation}
h_{final} = [h_{e1};h_{e2};|h_{e1} - h_{e2}|;h_{graph}]
\end{equation}
\begin{equation}
p = Sigmode(w^{T}h_{final} + b)
\end{equation}

\subsection{Contrastive Learning}
To facilitate the capturing of shared features and the differentiation of non-shared features between original and complement text, we propose a semi-supervised contrastive learning strategy\cite{khosla2020supervised} for integrating the enhanced layers. We consider the encoded vectors of original and corresponding complement sentences as positive examples, since they are semantically very close. Moreover, to obtain more negative examples, we regard other complement sentences in the same batch as negative examples for the original sentence. By optimizing the contrastive learning, the model will reduce the distance between the encoded vectors of original and corresponding complement sentences while enlarging the distance between the encoded vectors of original and other unrelated sentences. As a result, the final model can better differentiate the shared semantic parts between original and complement sentences.

To be more specific, firstly, we obtain the encoded vectors of the original sentences and the complement sentences for conducting contrastive learning, respectively.
\begin{equation}
\hat{h}_1 = FNN(h_1)
\end{equation}
\begin{equation}
\hat{h}_2 = FNN(h_2)
\end{equation}
\begin{equation}
\hat{h}^{'}_1 = FNN(h^{\prime}_1)
\end{equation}
\begin{equation}
\hat{h}^{'}_2 = FNN(h^{\prime}_2)
\end{equation}

The training objective is to minimize the semantic distance between the original sentence and the complement sentence, while maximizing the distance between unrelated sentences. We utilize InfoNCE loss\cite{oord2018representation} as the training loss and cosine similarity to measure the semantic distance between sentences.
\begin{equation}
\mathcal{L}_{\text{contrast1}}=-\sum_{i=1}^{N}\log \frac{e^{\operatorname{sim}\left(\hat{h}_{i,1}, \hat{h}_{i,1}^{\prime} \right) / \tau}}{\sum_{j=1}^{BatchSize} e^{\operatorname{sim}\left(\hat{h}_{i,1}, h_{j,1}^{'}\right) / \tau}}
\end{equation}

\begin{equation}
\mathcal{L}_{\text {contrast2}}=-\sum_{i=1}^{N}\log \frac{e^{\operatorname{sim}\left(\hat{h}_{i,2}, \hat{h}_{i,2}^{\prime} \right) / \tau}}{\sum_{j=1}^{BatchSize} e^{\operatorname{sim}\left(\hat{h}_{i,2}, h_{j,2}^{'}\right) / \tau}}
\end{equation}

where $N$ is the total number of samples. $h_{j,k}^{'}$ represents the $kth$ original sentence of the $jth$ sample in the current batch. $\tau$ is the temperature hyperparameter, which is used to control the model's differentiation of negative samples.

\subsection{Loss Function}
We apply the BCE loss to optimize the sentence matching task during the training process.
\begin{equation}
\mathcal{L}_{\text {binary}}=-\sum_{i=1}^{N}y_{i}log(p_i) + (1-y_i)log(1-p_i)
\end{equation}
The final total training loss function:
\begin{equation}
\mathcal{L} = \alpha\mathcal{L}_{\text {binary}}
+ \beta\mathcal{L}_{\text {contrast1}} + \gamma\mathcal{L}_{\text {contrast2}}
\end{equation}
$\alpha$, $\beta$, and $\gamma$ are hyperparameters set to 0.8, 0.1, 0.1 respectively.

\subsection{Generating Complementary Sentences}
To generate high-quality complement sentences, we constructed a dataset for training the SimBERT model. We can generate complementary sentences for existing data through this model. The model improves the attention mask mechanism in Transformer\cite{vaswani2017attention}, resulting in efficient performance when generating similar sentences.

Specifically, during the data collection stage, we collected user query statements and corresponding advertisements returned by search engines (UC and Quark). For advertisements with low click-through rates, we considered them irrelevant to the user's search intent and query statement. Therefore, we filtered out ad texts below the threshold $K$, and paired ad texts above the threshold with their corresponding queries to form sentence pairs. During the training phase, we obtained a total of 30,000k sentence pairs for model training. Finally, in the testing phase, the trained SimBERT model was frozen and used to generate complement sentences for the original sentences in the public dataset.
% \paragraph{Paragraphs}

% There is also a \verb+\paragraph+ command available, which sets the heading in
% bold, flush left, and inline with the text, with the heading followed by 1\,em
% of space.

\section{ Experiments}
\label{others}

\subsection{Experimental Setings}
We applied Bert-base-chinese as our encoder and AdamW as the optimizer during the training phase. In addition, we implemented a hierarchical learning rate strategy, with a learning rate of 1e-4 for the pre-training model and 2e-5 for the other parts. The model was trained for 30 epochs on a GPU server with a 3080 GPU, using a batch size of 16. Early stopping was employed.

\subsection{Dataset}
To validate the effectiveness of the proposed KSTM model, we conducted experiments on two publicly available Chinese datasets, i.e.,BQ\cite{chen2018bq} and LCQMC\cite{liu2018lcqmc}.

\textbf{Bank Question(BQ):} This is a large-scale question matching dataset related to the banking and finance domain, which has been widely used in sentence matching tasks. It contains a total of 120k sentence pairs, with the training, validation, and test sets containing 100k, 10k, and 10k sentence pairs, respectively.

\textbf{Large-scale Chinese Question Matching Corpus (LCQMC):} This is a semantic matching dataset of questions collected from the Baidu Knowledge website. The dataset focuses on matching intentions rather than meanings. It contains a total of 260k sentence pairs, with the training, validation, and test sets containing 239k, 8.4k, and 12.5k sentence pairs, respectively.

\subsection{Metrics}
Compared to previous research, we evaluated our approach on two publicly available datasets using the same evaluation metrics as previous studies, i.e., $ACC$ and $F1$. The calculation formula of $ACC$ and $F1$ are as follows,
\begin{equation}
ACC = \frac{TP + TN}{TP + TN + FP + FN}
\end{equation}
\begin{equation}
P = \frac{TP}{TP + FP}
\end{equation}
\begin{equation}
R = \frac{TP}{TP + FN}
\end{equation}
\begin{equation}
F1 = \frac{2 * P * R}{P + R}
\end{equation}
where $TP$ and $FP$ respectively represent the number of cases correctly predicted as positive and the number of cases incorrectly predicted as positive. $TN$ and $FN$ respectively represent the number of cases correctly predicted as negative and the number of cases incorrectly predicted as negative.

\subsection{Baselines}
To demonstrate the effectiveness of our approach, we compared KSTM model with the following strong baseline models, which use BERT as encoder.

\textbf{Glyce\cite{meng2019glyce}:} A glyph-based model for Chinese NLP tasks.

\textbf{GMN\cite{chen2020neural}:} A sentence matching framework capable of dealing with multi-granular input information.

\textbf{LET\cite{lyu2021let}:}  A short Text Matching model using external knowledge to eliminate sentence ambiguity.

\textbf{DSSTM\cite{deng2022enhanced}}: Combining the relative distance information of words, a distance-aware self-attention and multi-level matching model is proposed to solve the Chinese short Text Matching task.

\textbf{OTE\cite{ma2022ote}:} A short Text Matching model based on external knowledge, combined with SoftLexico model and  hybrid data augmentation to obtain multi-granularity features.

\textbf{CBM\cite{chen2022context}:} This method proposes to use external knowledge to enhance the semantic information of the original short text.

\subsection{Experiment Result}

We compared our KSTM model with other short Text Matching models that use pre-trained Bert models as encoders on two datasets, using $F1$ and $accuracy$ as evaluation metrics. As shown in Table 1, the KSTM model outperforms all baseline models on both datasets. On the BQ dataset, the KSTM model achieves 1.46\% higher accuracy and 1\% higher $F1$ than the present SOTA model CBM. On the LCQMC dataset, the KSTM model outperforms the present best model DSSTM by 0.1\% in accuracy and 1.1\% in $F1$. This demonstrates the effectiveness of our proposed KSTM model.

\begin{table}[h]
\caption{Performance comparison results of KSTM and baseline models on BQ and LCQMC datasets.}
\centering
\begin{tabular}{@{}ccccccc@{}}
\toprule
\multirow{2}{*}{\textbf{Model}} &  & \multicolumn{2}{c}{\textbf{BQ}} &  & \multicolumn{2}{c}{\textbf{LCQMC}} \\ \cmidrule(l){3-7} 
                                &  & ACC            & F1             &  & ACC              & F1              \\ \midrule
Glyce                           &  & 85.80          & 85.50          &  & 88.70            & 88.80           \\
LET                             &  & 85.30          & 84.98          &  & 88.38            & 88.85           \\
GMN                             &  & 85.60           & 85.50          &  & 87.30            & 88.00           \\
DSSTM                           &  & 85.40          & -            &  & 88.90            & -              \\
OTE                             &  & 85.26          & 84.77          &  & 86.68            & 88.29           \\
CBM                             &  & 86.16          & 87.44          &  & 88.80             & 89.10            \\
\textbf{KSTM(our)}               &  & \textbf{87.62}          & \textbf{88.44}          &  &    \textbf{89.00}              & \textbf{90.20}                \\ \bottomrule
\end{tabular}
\end{table}

\subsection{Ablation Experiment}
% \usepackage{multirow}
\begin{table}[h]
\caption{Ablation experiment results of the KSTM model.}
\centering
\begin{tabular}{ccccccc}
\hline
\multirow{2}{*}{\textbf{Model}} &  & \multicolumn{2}{c}{\textbf{BQ}} &  & \multicolumn{2}{c}{\textbf{LCQMC}} \\ \cline{3-7} 
                                &  & ACC            & F1             &  & ACC              & F1              \\ \hline
w/o Contrastive Learning        &  & 86.10           & 87.50          &  & 88.10            & 88.80           \\
w/o Knowledge Graphs            &  & 85.30          & 85.88          &  & 85.38            & 86.85           \\
w/o Complementary Sentence      &  & 85.68          & 85.50          &  & 84.30            & 86.10           \\
\textbf{KSTM(our)}              &  & \textbf{87.62} & \textbf{88.44} &  & \textbf{89.00}    & \textbf{90.20}   \\ \hline
\end{tabular}
\end{table}

In order to demonstrate the effectiveness of the components of KSTM, we conducted three ablation experiments on the BQ and LCQMC datasets, as shown in Table 2. The first experiment aimed to verify the effectiveness of contrastive learning. We removed contrastive learning between the original sentence and its complementary sentence and used only binary cross-entropy as the overall loss function. The second experiment aimed to demonstrate the improvement in model performance by introducing external knowledge into the model. We did not use external knowledge graphs or graph encoding information, only enhanced sentence encoding for sentence matching. The third experiment aimed to demonstrate the effectiveness of the complementary sentences we constructed. Following the unsupervised contrastive learning approach proposed by SimCES\cite{gao2021simcse}, we replaced the complementary sentences with the original sentences, we inputted the original sentences into the model to obtain the encodings, which were not identical due to the presence of dropout. Two identical original sentences were used as positive examples for contrastive learning.

The experimental results demonstrate that removing any part of the model will have a negative impact on its performance. Specifically, when we remove contrastive learning, the model's accuracy and F1 score on the BQ dataset decrease by 1.52\% and 0.94\%, respectively. On the FCQMC dataset, the accuracy and F1 score decrease by 0.9\% and 1.4\%, respectively. When we remove the knowledge graph, the model's accuracy and F1 score on the BQ dataset decrease by 2.32\% and 2.56\%, respectively. On the FCQMC dataset, the accuracy and F1 score decrease by 3.62\% and 3.35\%, respectively. When we remove the use of complementary sentences and use unsupervised contrastive learning instead, the model's accuracy and F1 score on the BQ dataset decrease by 1.94\% and 2.94\%, respectively. On the FCQMC dataset, the accuracy and F1 score decrease by 4.7\% and 4.1\%, respectively.

Through comparing three ablation strategies, we found that compared to external knowledge, removing contrastive learning has a smaller impact on the model. We believe this is because the role of contrastive learning is to guide the model to generate more semantically meaningful encoding. It can guide the model to use semantic vector representations to represent text with similar semantics using vectors that are close in space, and use vectors that are far apart in space for text with different semantics. For short-text datasets, pre-trained models already have strong encoding capabilities to understand and encode their semantic information, so the effect of using contrastive learning to improve model performance is lower. In addition, the use of complementary sentences and knowledge graphs has a significant positive effect on model performance. This is because they both introduce external knowledge, which largely solves the problem of the model lacking sufficient information to judge whether two sentences match in short-text scenarios. Comparing the results of the two datasets, removing knowledge graphs or complementary sentences results in a greater performance decrease on the LCQMC dataset than on the BQ dataset. We believe this is because the sentence matching in the LCQMC dataset is more focused on sentence intent rather than meaning, which requires the model to have more information to understand the sentence and perform intent inference. Therefore, the method of using external knowledge has a more significant improvement effect on models trained on this dataset.

\section{Conclusion}
In this paper, we propose a model(KSTM) that combines external knowledge and contrastive learning to address the problem of insufficient semantic information in short Text Matching tasks. We use two methods to introduce external knowledge, one of which is to generate complement sentences using the SimBert model, and the other is to construct a graph using keywords and knowledge words. Additionally, we use contrastive learning to improve the model's ability to encode sentence representations. We validate the KSTM model on two Chinese short Text Matching tasks, and the results show that the KSTM model outperforms the previous baseline model. In future work, we plan to introduce more powerful models such as ChatGPT to generate more accurate complement sentences. Additionally, in constructing the knowledge graph, we will design a refined filtering mechanism to filter out knowledge words retrieved from the knowledge base that are unrelated to the keywords, thereby reducing the influence of noise.

\bibliographystyle{unsrt}

\end{document}